\documentclass[letterpaper]{article} 
\usepackage{aaai24}  
\usepackage{times}  
\usepackage{helvet}  
\usepackage{courier}  
\usepackage[hyphens]{url}  
\usepackage{graphicx} 
\usepackage{natbib}  
\usepackage{caption} 
\usepackage{algorithm}
\usepackage{algorithmic}

\usepackage{newfloat}
\usepackage{listings}

\usepackage{url}            
\usepackage{booktabs}       
\usepackage{amsfonts}       
\usepackage{nicefrac}       
\usepackage{microtype}      
\usepackage[table]{xcolor}        

\usepackage{booktabs}       
\usepackage{amssymb}
\usepackage{nicefrac}       
\usepackage{amsmath}
\usepackage{xspace}
\usepackage{multirow}
\usepackage{booktabs}
\usepackage{color}
\usepackage{graphicx}
\usepackage{colortbl}
\usepackage{subcaption}
\usepackage{adjustbox}

\DeclareCaptionStyle{ruled}{labelfont=normalfont,labelsep=colon,strut=off} 
\floatstyle{ruled}
\newfloat{listing}{tb}{lst}{}
\floatname{listing}{Listing}
\makeatletter
\DeclareRobustCommand\onedot{\futurelet\@let@token\@onedot}
\def\@onedot{\ifx\@let@token.\else.\null\fi\xspace}

\def\eg{\emph{e.g}\onedot} 
\def\ie{\emph{i.e}\onedot}

\def\etal{\emph{et al}\onedot}
\makeatother

\title{Simple Image-level Classification Improves Open-vocabulary Object Detection}
\author{
    Ruohuan Fang\textsuperscript{\rm 1},
    Guansong Pang\textsuperscript{\rm {2*}},
    Xiao Bai\textsuperscript{\rm 1,\rm 3}\thanks{Corresponding authors: 
    G. Pang (gspang@smu.edu.sg) and 
    X. Bai (baixiao@buaa.edu.cn)}
}
\affiliations{
    \textsuperscript{\rm 1}School of Computer Science and Engineering, Beihang University\\
    \textsuperscript{\rm 2}School of Computing and Information Systems, Singapore Management University\\
    \textsuperscript{\rm 3}State Key Laboratory of Software Development Environment, Jiangxi Research Institute, Beihang University\\ 
}

\usepackage{bibentry}

\begin{document}

\maketitle

\begin{abstract}
Open-Vocabulary Object Detection (OVOD) aims to detect novel objects beyond a given set of base categories on which the detection model is trained. Recent OVOD methods focus on adapting the image-level pre-trained vision-language models (VLMs), such as CLIP, to a region-level object detection task via, \eg, region-level knowledge distillation, regional prompt learning, or region-text pre-training, to expand the detection vocabulary. These methods have demonstrated remarkable performance in recognizing regional visual concepts, but they are weak in exploiting the VLMs' powerful global scene understanding ability learned from the billion-scale image-level text descriptions. This limits their capability in detecting hard objects of small, blurred, or occluded appearance from novel/base categories, whose detection heavily relies on contextual information. To address this, we propose a novel approach, namely \textbf{S}imple \textbf{I}mage-level \textbf{C}lassification for \textbf{C}ontext-\textbf{A}ware \textbf{D}etection \textbf{S}coring (SIC-CADS), to leverage the superior global knowledge yielded from CLIP for complementing the current OVOD models from a global perspective. The core of SIC-CADS is a multi-modal multi-label recognition (MLR) module that learns the object co-occurrence-based contextual information from CLIP to recognize all possible object categories in the scene. These image-level MLR scores can then be utilized to refine the instance-level detection scores of the current OVOD models in detecting those hard objects. 
This is verified by extensive empirical results on two popular benchmarks, OV-LVIS and OV-COCO, which show that SIC-CADS achieves significant and consistent improvement when combined with different types of OVOD models. Further, SIC-CADS also improves the cross-dataset generalization ability on Objects365 and OpenImages. Code is available at \url{https://github.com/mala-lab/SIC-CADS}.
\end{abstract}

\begin{figure}[t]
        \centering
        \includegraphics[width=0.47\textwidth]{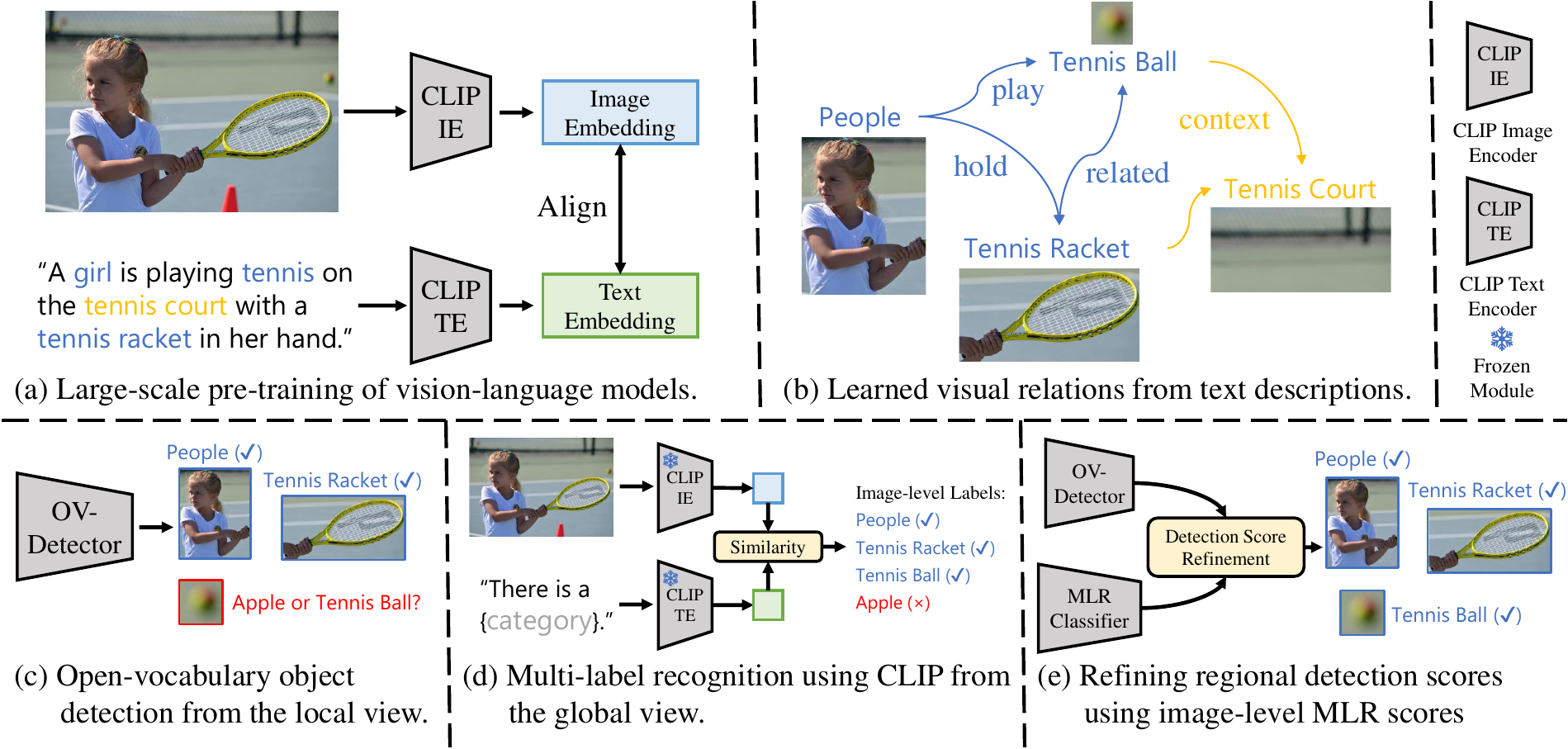}
        \caption{\textbf{Motivation of our method. } (\textbf{a}) Large VLMs like CLIP~\cite{radford2021learning} can understand the image globally by learning rich knowledge from a huge amount of image-text pairs. Such knowledge can include diverse relations between visual concepts in the scene, as shown in (\textbf{b}). As illustrated in (\textbf{c}), current OVOD approaches focus on regional visual concept detection but they are weak in exploiting this global knowledge, which can fail to detect novel hard objects, such as 
        ambiguous objects like the blurred tennis ball. (\textbf{d}) We instead learn an image-level multi-label recognition (MLR) module to leverage the global knowledge yielded from CLIP for recognizing those hard objects. (\textbf{e}) The image-level MLR scores are then utilized to refine the instance-level detection scores from a global perspective for more effective OVOD.
        }
        \label{fig:motivation}
\end{figure}

\section{Introduction}
Open-Vocabulary Object Detection (OVOD) is a challenging task that requires detecting objects of novel categories that are not present in the training data. To tackle this problem, conventional approaches focus on leveraging external image-text data as weak supervisory information to expand the detection vocabulary of the categories. In recent years, large-scale pre-trained vision-language models (VLMs), \eg, CLIP~\cite{radford2021learning} and ALIGN~\cite{jia2021scaling}, which are trained using billion-scale internet-crawled image-caption pairs (see Fig. \ref{fig:motivation}(a) for an example), have been widely used to empower OVOD. 

Existing VLM-based OVOD studies focus on how to adapt the image-level pre-trained CLIP to a region-level object detection task. Typically they adopt a regional concept learning method, such as Region-level Knowledge Distillation~\cite{du2022learning, bangalath2022bridging} that aligns region embeddings to their corresponding features extracted from the image encoder of CLIP, Regional Prompt Learning~\cite{wu2023cora, du2022learning, feng2022promptdet} that learns continuous prompt representations to better pair with region-level visual embeddings, Region-Text Pre-training~\cite{zhong2022regionclip} that explicitly aligns image regions and text tokens during vision-language pre-training, or Self-Training~\cite{zhou2022detecting} that generates pseudo-labels of novel objects on the image-level labeled datasets (\eg, ImageNet-21k~\cite{deng2009imagenet} and Conceptual Captions~\cite{sharma2018conceptual}). 

These methods have demonstrated remarkable performance in recognizing regional visual concepts, but they are weak in exploiting the VLMs' powerful global scene understanding ability that can capture important relations between different visual concepts \cite{wu2023aligning, zhong2021learning}. These relations can be the co-occurrence of different objects such as the tennis ball and the tennis racket, or their interdependence to commonly shared background environment such as the tennis ball and tennis court, as shown in Fig. \ref{fig:motivation}(b). This weakness can limit their capability in detecting \textbf{hard objects} that are of small, blurred, or occluded appearance, whose detection relies heavily on contextual features of other objects in the same image. Further, the OV detectors have naturally learned the context information for base categories when aligning regional features with corresponding text embeddings, since the network can automatically capture the context features within the receptive field that are related to base objects. However, the context features related to novel objects are not learned due to the absence of novel object annotations during the training of OV detectors. This can be one of the key reasons that incurs the performance gap between base and novel objects, especially the novel hard objects. 

To address this issue, this work instead aims to utilize the global scene understanding ability for OVOD. The key motivation is that the image-level embeddings extracted from CLIP's image encoder carry global features about various objects in the entire scene, which are semantically related in the natural language descriptions. This knowledge can then provide important contextual information for detecting the aforementioned hard objects, \eg, the small, blurred tennis ball in Fig. \ref{fig:motivation}(c), which are otherwise difficult to detect using only regional features. 

Inspired by this, we propose a novel approach that utilizes a \textbf{S}imple \textbf{I}mage-level \textbf{C}lassification module for \textbf{C}ontext-\textbf{A}ware \textbf{D}etection \textbf{S}coring, termed SIC-CADS. Our image-level classification task is specified by a Multi-Label Recognition (MLR) module which learns multi-modal knowledge extracted from CLIP. The MLR module predicts image-level scores of different possible object categories that could exist in a specific scene. For example, as shown in Fig. \ref{fig:motivation}(d), context information of the tennis racket and tennis court helps recognize the blurred tennis ball in such a sports-related scene. Thus, the image-level MLR scores can be used to refine the instance-level detection scores of existing OVOD models for improving their detection performance from a global perspective, as shown in Fig. \ref{fig:motivation}(e). 

Our main contributions are summarized as follows. (i) We propose a novel approach SIC-CADS that utilizes a MLR module to leverage VLMs' global scene knowledge for improving OVOD performance. (ii) SIC-CADS is a simple, lightweight, and generic framework that can be easily plugged into different existing OVOD models to enhance their ability to detect hard objects. (iii) Extensive experiments on OV-LVIS, OV-COCO, and cross-dataset generalization benchmarks show that SIC-CADS significantly boosts the detection performance when combined with different types of state-of-the-art (SOTA) OVOD models, achieving 1.4 - 3.9 gains of AP$_{r}$ for OV-LVIS and 1.7 - 3.2 gains of AP$_{novel}$ for OV-COCO. Besides, our method also largely improves their cross-dataset generalization ability, yielding 1.9 - 2.1 gains of mAP$_{50}$ on Objects365~\cite{shao2019objects365} and 1.5 - 3.9 gains of mAP$_{50}$ on OpenImages~\cite{kuznetsova2020open}. 

\begin{figure*}[t]
        \centering
        \includegraphics[width=0.80\textwidth]{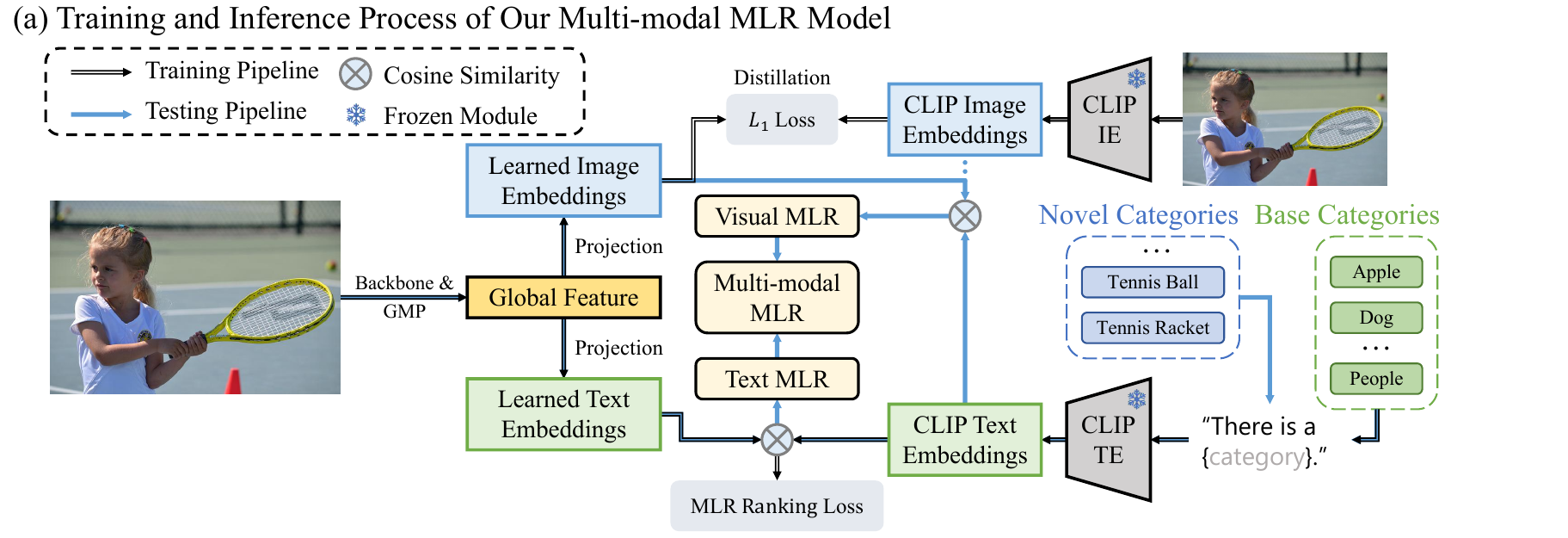}
        \includegraphics[width=0.80\textwidth]{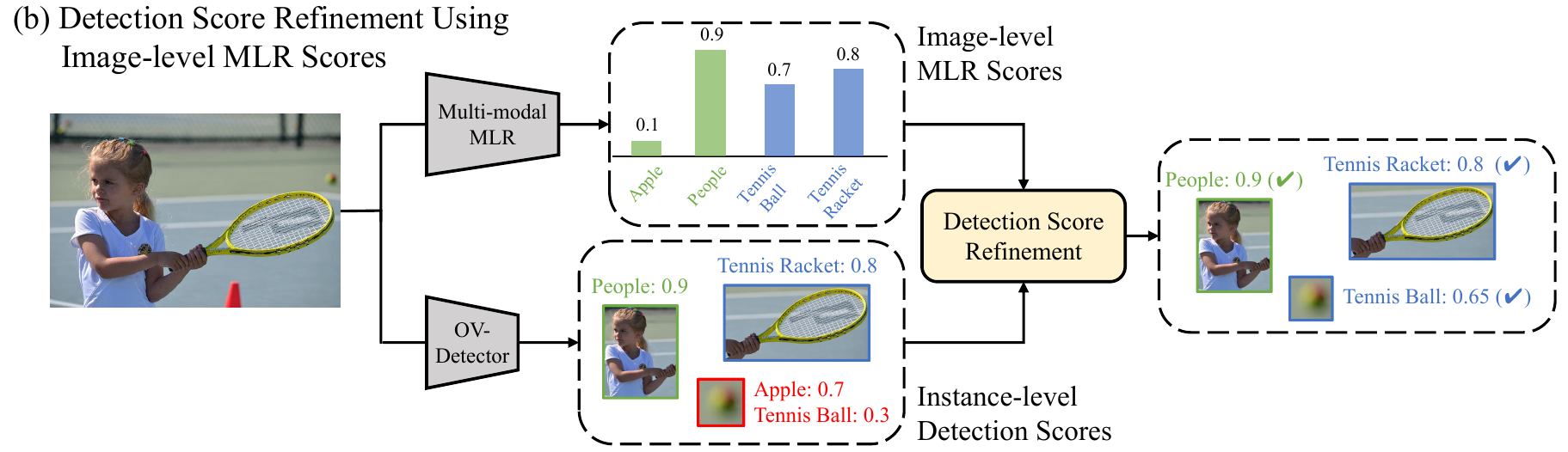}
        \caption{\textbf{Overview of our approach SIC-CADS.} \textbf{(a)} During training, our proposed MLR module learns CLIP's global multi-modal knowledge from the text encoder (text MLR) and image encoder (visual MLR). During inference, the two branches demonstrate superior performance in recognizing objects from base and novel classes respectively. Hence, we combine the two branches to make our full model, multi-modal MLR. 
        \textbf{(b)} Our MLR module can be plugged into existing OVOD models, via a simple detection score refinement process, to boost the performance in detecting hard objects from a global perspective.}
        \label{fig:overview}
\end{figure*}

\section{Related Work}
\textbf{Vision-Language Models (VLMs):} The task of vision-language pre-training is to learn aligned multimodal representations from image-text datasets. Early works~\cite{joulin2016learning, li2017learning} demonstrate that CNNs trained to predict words in image captions can learn representations competitive with ImageNet training. Recently, the large-scale VLMs (\eg, CLIP~\cite{radford2021learning} and ALIGN~\cite{jia2021scaling}) which are contrastively pre-trained on billion-scale internet-sourced image-caption datasets, have shown powerful zero-shot performance on image classification tasks. 
With the advent of large VLMs, they have been applied to various downstream vision tasks, including object detection~\cite{gu2021open, du2022learning, zhou2022detecting}, instance segmentation~\cite{xu2022groupvit, ghiasi2021open}, image generation~\cite{nichol2021glide, ramesh2021zero}, and anomaly detection~\cite{zhou2023anomalyclip,wu2023vadclip}.  

\noindent\textbf{Open-Vocabulary Object Detection (OVOD):} Traditional object detection models can only detect the base categories which are presented during training. OVOD aims to extend the vocabulary of object detectors using additional large image-text datasets. OVR-CNN~\cite{zareian2021open} first formulates this problem and proposes its baseline method by aligning the regional features with words in the caption. ViLD~\cite{gu2021open} addresses the problem of OVOD by distilling the regional representations using CLIPs' image encoder. Detic~\cite{zhou2022detecting} adopts self-training which produces pseudo-labels of novel objects on the datasets such as ImageNet-21k~\cite{deng2009imagenet} to expand the detection vocabulary. DetPro~\cite{du2022learning}, PromptDet~\cite{feng2022promptdet}, and POMP~\cite{ren2023prompt} use prompt tuning to adapt the image-level pre-trained VLMs to the region-level object detection task. RegionCLIP~\cite{zhong2022regionclip} learns region-level visual representations by region-text pre-training. F-VLM~\cite{kuo2022f} directly utilizes the regional features of frozen VLMs for object recognition and eliminates the need for regional knowledge distillation. Our simple MLR network adopts a similar score fusion strategy as ViLD and F-VLM. The key difference is that we recognize all objects via multi-modal MLR, while ViLD and F-VLM use an ensembled score for region-level object classification. Plus, our method is also orthogonal to ViLD and F-VLM, as shown in Tab. \ref{tab::ov_lvis} that our SIC-CADS can be plugged into ViLD to largely improve its performance.

\noindent\textbf{Zero-Shot Multi-Label Recognition (ZS-MLR):} The task of multi-label recognition is to predict the labels of all objects in one image. As an extension of Zero-Shot Learning (ZSL)~\cite{romera2015embarrassingly, xian2017zero, xian2019f, xu2020attribute}, ZS-MLR aims to recognize both seen and unseen objects in the image. The key of ZS-MLR is to align the image embeddings with the category embeddings. Previous ZS-MLR methods~\cite{zhang2016fast, ben2021semantic, huynh2020shared, narayan2021discriminative, gupta2021generative} mostly use the single-modal embeddings from language models (\eg, GloVe~\cite{pennington2014glove}) and adopt different strategies (\eg, attention mechanism~\cite{narayan2021discriminative, huynh2020shared} or generative models~\cite{gupta2021generative}) to boost the performance. He \etal~\cite{he2022open} further propose Open-Vocabulary Multi-Label Recognition (OV-MLR) task which adopts the multi-modal embeddings from CLIP for MLR. 
They use the transformer backbone to extract regional and global image representations and a two-stream module to transfer knowledge from CLIP. Recognize Anything Model (RAM)\cite{zhang2023recognize} leverages the large-scale image-text pairs by parsing the captions for automatic image labeling. Then, they train an image-tag recognition decoder which reveals strong ZS-MLR performance.

\section{Method}
This work aims to exploit CLIP's image-level/global knowledge to better recognize hard objects for more effective OVOD. We first train a multi-label recognition (MLR) module by transferring global multi-modal knowledge from CLIP to recognize all existing categories in the entire scene. Then during inference, our MLR module can be easily plugged into existing OVOD models, via a simple detection score refinement process, to boost the detection performance. 

\begin{table*}
  \begin{minipage}{0.68\linewidth}
    \centering
    \caption{Enabling different SOTA OVOD models on the OV-LVIS benchmark. }\label{tab::ov_lvis}
    \adjustbox{width=1.0\textwidth}{
        \begin{tabular}{{l|c|c|>{\columncolor{gray!20}}cccc}}
        \toprule
        Method&Backbone &Supervision  &AP$_{r}$&AP$_{c}$&AP$_{f}$ & AP   \\\midrule
        BoxSup~\cite{zhou2022detecting}&\multirow{2}*{\shortstack{CenterNet-V2 + \\ ResNet50}} &\multirow{2}*{\shortstack{LVIS-Base \\+ CLIP }}&16.4&31.0&35.4&30.2 \\
        \hspace{0.4cm} + SIC-CADS& & &20.3\textcolor{blue}{(+3.9)}&31.8\textcolor{blue}{(+0.8)}&35.5\textcolor{blue}{(+0.1)}&31.2\textcolor{blue}{(+1.0)} \\ \midrule

        ViLD~\cite{gu2021open}&\multirow{2}*{\shortstack{Faster-RCNN + \\ ResNet50}} &\multirow{2}*{\shortstack{LVIS-Base \\+ CLIP }}&16.8&25.6&28.5&25.2 \\
        \hspace{0.4cm} + SIC-CADS & & &18.7\textcolor{blue}{(+2.1)}& 26.4\textcolor{blue}{(+0.8)}&28.7\textcolor{blue}{(+0.2)}&26.1\textcolor{blue}{(+0.9)} \\ \midrule
        
        RegionCLIP~\cite{zhong2022regionclip}&\multirow{2}*{\shortstack{Faster-RCNN + \\ ResNet50×4}} &\multirow{2}*{\shortstack{LVIS-Base + \\CLIP + CC3M~\cite{sharma2018conceptual} }} &19.7&28.2&30.7&27.7 \\
        \hspace{0.4cm} + SIC-CADS& & &21.9\textcolor{blue}{(+2.2)}&29.1\textcolor{blue}{(+0.9)}&30.9\textcolor{blue}{(+0.2)}&28.5\textcolor{blue}{(+0.8)} \\ \midrule
        OC-OVD~\cite{bangalath2022bridging}&\multirow{2}*{\shortstack{Faster-RCNN + \\ ResNet50}} &\multirow{2}{*}{\shortstack{LVIS-Base + CLIP + \\LMDet~\cite{maaz2021multi} + IN21k}} &21.1&25.0&29.1&25.9 \\
        \hspace{0.4cm} + SIC-CADS & & &22.5\textcolor{blue}{(+1.4)}&25.6\textcolor{blue}{(+0.6)}&29.0\textcolor{red}{(-0.1)}&26.3\textcolor{blue}{(+0.4)} \\ \midrule
        Detic~\cite{zhou2022detecting}&\multirow{2}*{\shortstack{CenterNet-V2 + \\ ResNet50}} &\multirow{2}{*}{\shortstack{LVIS-Base +  \\CLIP + IN21k~\cite{deng2009imagenet}}} &24.9&32.5&35.6&32.4 \\
        \hspace{0.4cm} + SIC-CADS & & &26.5\textcolor{blue}{(+1.6)}&33.0\textcolor{blue}{(+0.5)}&35.6\textcolor{gray}{(+0.0)}&32.9\textcolor{blue}{(+0.5)} \\ \midrule
        POMP~\cite{ren2023prompt}&\multirow{2}*{\shortstack{CenterNet-V2 + \\ ResNet50}} &\multirow{2}*{\shortstack{LVIS-Base +  \\CLIP + IN21k~\cite{deng2009imagenet}}} &25.2&33.0&35.6&32.7 \\
        \hspace{0.4cm} + SIC-CADS&& &26.6\textcolor{blue}{(+1.4)}&33.3\textcolor{blue}{(+0.3)}&35.6\textcolor{gray}{(+0.0)}&33.1\textcolor{blue}{(+0.4)} \\ \bottomrule
        \end{tabular}
    }
  \end{minipage}
  \hfill
  \begin{minipage}{0.3\linewidth}
    \centering
    \caption{Results on the cross-dataset evaluation. }
    \label{tab::cross_dataset}
    \adjustbox{width=1.0\textwidth}{
        \begin{tabular}{l|cc}\toprule
        Method & Objects365 & OpenImages\\\midrule
        BoxSup&26.6&46.4\\
        \hspace{0.4cm} + SIC-CADS&28.7\textcolor{blue}{(+2.1)}&50.3\textcolor{blue}{(+3.9)} \\ \midrule
        Detic&29.3&53.2\\
        \hspace{0.4cm} + SIC-CADS&31.2\textcolor{blue}{(+1.9)}&54.7\textcolor{blue}{(+1.5)} \\ \bottomrule
        \end{tabular}
    }
  \end{minipage}
\end{table*}

\subsection{Preliminaries\label{sec::pre}}
In OVOD, typically we have an object detector trained with a detection dataset $\mathcal{D}_{\mathrm{det}}$ which contains the exhaustively annotated bounding-box labels for a set of base categories $\mathcal{C}_\text{B}$. Some external image-caption datasets may be available that can be used to expand the detection vocabulary, enabling the detector to recognize novel categories beyond the closed set of base categories. 
During inference, the categories in the testing set comprise both base categories $\mathcal{C}_\text{B}$ and novel categories $\mathcal{C}_\text{N}$, \ie, $\mathcal{C}_{{test}}=\mathcal{C}_{\text{B}} \cup \mathcal{C}_{\text{N}}$ and $\emptyset =\mathcal{C}_{\text{B}} \cap \mathcal{C}_{\text{N}}$. 

Therefore, the OVOD models are required to solve two subsequent problems: (1) the effective localization of all objects in the image, and (2) the correct classification of each object into one of the class labels in $\mathcal{C}_{{test}}$. 
Many VLM-based OVOD methods~\cite{zhou2022detecting, gu2021open, bangalath2022bridging} adopt the two-stage detection framework (\eg, Mask-RCNN~\cite{he2017mask}) as the base detector. In the first stage, a RPN (\emph{region proposal network}) takes an image $\mathbf{I} \in \mathbb{R}^{H \times W \times 3}$ as input, and produces a set of class-agnostic object proposals $\mathcal{P} \subset \mathbb{R}^{4}$ which denotes the coordinates for the proposal boxes. In the second stage, a RoI (\emph{region of interest}) Align head computes the pooled representations $\mathcal{E} = \{e_r\}_{r \in \mathcal{P}} \subset \mathbb{R}^{d}$ for the proposals $\mathcal{P}$. To classify each proposal into one of the categories in $\mathcal{C}_{\text{B}} \cup \mathcal{C}_{\text{N}}$, for each category $c$, we obtain its text embedding $t_c$ by feeding the category name in a prompt template, \eg, \texttt{`a photo of a \{category name\}'} into the text encoder $\mathcal{T}$ of a pre-trained VLM like CLIP. The probability of a region proposal $r$ being classified into category $c$ is computed as:
\begin{equation}
  p(r, c) = \frac
  {\exp\left(\mathit{cos}(e_r, t_c)/\tau\right)}
  {\sum\limits_{c' \in \mathcal{C}_{test} } \exp\left(\mathit{cos}(e_r, t_{c'})/\tau\right)},
\end{equation}
where $\mathit{cos}(\cdot,\cdot)$ denotes a cosine similarity and $\tau$ is a temperature scaling factor.

\subsection{CLIP-driven Multi-modal MLR Modeling \label{sec::mlr}}

\textbf{Overall Framework:} Our proposed MLR module is designed to leverage the superior scene understanding ability of CLIP for effective detection of both novel and base categories. Specifically, as shown in Fig. \ref{fig:overview}(a), given an input image $\mathbf{I} \in \mathbb{R}^{H \times W \times 3}$, we first extract the multi-scale feature maps $\mathcal{P} = \{P_2,P_3,P_4,P_5\}$ using a ResNet50-FPN backbone, followed by a Global Max Pooling (GMP) operator and a concatenation of all feature vectors in different FPN levels to obtain a global image embedding $e^{\mathit{global}}$. The global image embedding is then utilized in two branches, including one \textbf{text MLR} branch which is aligned $e^{\mathit{global}}$ to the text embeddings of different categories yielded from the text encoder of CLIP, and one \textbf{visual MLR} branch which distills the global image embedding from the image encoder of CLIP into $e^{\mathit{global}}$. During inference, the text MLR is weaker in recognizing novel categories since it is trained with only base categories, while the visual MLR can recognize both base and novel categories by distilling the zero-shot recognition ability of CLIP. Therefore, we combine the scores of both branches to achieve better recognition performance for novel and base categories, which is noted as \textbf{Multi-modal MLR}. Below we introduce them in detail.

\noindent\textbf{Text MLR:} This branch aims to align the image embeddings with the corresponding category text embeddings yielded from the text encoder of CLIP so that each image can be classified based on CLIP text embeddings. We first use a linear layer $f(\cdot)$ to project the image embedding $e^{\mathit{global}}$ to a new feature space, namely learned text embedding space, and obtain $e^{text} = f(e^{\mathit{global}}) \in \mathbb{R}^{d}$. 
We then use the commonly used prompt template \texttt{`there is a \{category\}'} and feed the template filled with category names into the CLIP text encoder to obtain category-specific text embeddings $t$. The classification score of each category $c$ for an image $i$ is computed using cosine similarity: 
\begin{equation}
  s_{i,c}^{text} = \mathit{cos}(e^{text}_i, t_{c}).
\end{equation}
To train this branch, we adopt a powerful MLR ranking loss to increase the scores of positive categories (\ie, categories that appear in the image) and decrease the scores of negative categories (\ie, categories that this image does not contain). Specifically, we define the rank loss as:
\begin{equation}\label{equ::rank}
  L_{rank} = \sum\limits_{i} \sum\limits_{p \in \mathcal{N}(i), n \notin \mathcal{N}(i)} \max(1 + s_{i,n}^{\mathit{text}} - s_{i,p}^{\mathit{text}}, 0),
\end{equation} 
where $\mathcal{N}(i)$ denotes the image-level labels of image $i$,
and $s_{i,n}$ and $s_{i,p}$ denote the classification scores of positive and negative categories w.r.t. the image, respectively. During training, text MLR learns to align the text embeddings of multi-label base categories for each image by minimizing $L_{rank}$. For a test image $j$, the classification score w.r.t. a category $c$ is defined as $s_{j,c}^{text}$.  

\begin{table*}
  \begin{minipage}{0.70\linewidth}
    \centering
    \caption{Enabling different SOTA OVOD models on the OV-COCO benchmark. }\label{tab::ov_coco}
    \adjustbox{width=1.0\textwidth}{
        \begin{tabular}{l|c|c|>{\columncolor{gray!20}}ccc}
        \toprule
        Method &Backbone&Supervision  & AP$_{novel}$&AP$_{base}$& AP   \\\midrule
        Detic~\cite{zhou2022detecting}&\multirow{2}*{\shortstack{Faster-RCNN + \\ ResNet50}} &\multirow{2}*{\shortstack{COCO-Base + CLIP \\+ COCO Captions~\cite{chen2015microsoft}}}&27.8&51.1&45.0 \\
        \hspace{0.4cm} + SIC-CADS & & & 31.0\textcolor{blue}{(+3.2)}&52.4\textcolor{blue}{(+1.3)}&46.8\textcolor{blue}{(+1.7)} \\ \midrule
        BARON~\cite{wu2023aligning}&\multirow{2}*{\shortstack{Faster-RCNN + \\ ResNet50}} &\multirow{2}*{\shortstack{COCO-Base + CLIP \\+ COCO Captions~\cite{chen2015microsoft}}} &35.1&55.2&49.9 \\
        \hspace{0.4cm} + SIC-CADS & & &36.9\textcolor{blue}{(+1.8)}&56.1\textcolor{blue}{(+0.9)}&51.1\textcolor{blue}{(+1.2)} \\ \midrule
        RegionCLIP~\cite{zhong2022regionclip} &\multirow{2}*{\shortstack{Faster-RCNN + \\ ResNet50×4}}&\multirow{2}*{\shortstack{COCO-Base + \\CLIP + CC3M~\cite{sharma2018conceptual} }} &39.3&61.6&55.7 \\
        \hspace{0.4cm} + SIC-CADS& & &41.4\textcolor{blue}{(+2.1)}&61.7\textcolor{blue}{(+0.1)}&56.3\textcolor{blue}{(+0.6)}\\ \midrule
        OC-OVD~\cite{bangalath2022bridging}&\multirow{2}*{\shortstack{Faster-RCNN + \\ ResNet50}} &\multirow{2}{*}{\shortstack{CLIP + LMDet~\cite{maaz2021multi}\\ + COCO Captions~\cite{chen2015microsoft}}} &40.7&54.1&50.6 \\
        \hspace{0.4cm} + SIC-CADS & & &42.8\textcolor{blue}{(+2.1)}&55.1\textcolor{blue}{(+1.0)}&51.9\textcolor{blue}{(+1.3)} \\ \midrule
        CORA~\cite{wu2023cora}&\multirow{2}*{\shortstack{DAB-DETR + \\ ResNet50×4}} &\multirow{2}{*}{\shortstack{\shortstack{COCO-Base + CLIP \\+ COCO Captions~\cite{chen2015microsoft}}}} &41.6&44.7&43.9 \\
        \hspace{0.4cm} + SIC-CADS & & &43.3\textcolor{blue}{(+1.7)}&45.7\textcolor{blue}{(+1.0)}&45.1\textcolor{blue}{(+1.2)} \\ \bottomrule
        \end{tabular}
      }
  \end{minipage}
  \hfill
  \begin{minipage}{0.25\linewidth}
    \centering
    \caption{Effectiveness of hyperparameter $\gamma$. }
    \label{tab::gamma}
    \adjustbox{width=1.0\textwidth}{
    \begin{tabular}{c|cccc}\toprule
          $\gamma$ & AP$_{r}$&AP$_{c}$&AP$_{f}$&AP\\\midrule
          0.3&19.3&31.6&\textbf{35.5}&31.0 \\
          0.4&20.0&\textbf{31.8}&\textbf{35.5}&31.2 \\ 
          0.5&20.3&\textbf{31.8}&\textbf{35.5}&\textbf{31.3}\\
          0.6&20.6&31.6 &35.3&{31.1}\\ 
          0.7&{20.7}& 31.1 &34.8&30.7\\ 
          0.8&\textbf{21.1}&  29.8 &33.9&30.8 \\ \bottomrule
          \end{tabular}
    }
  \end{minipage}
\end{table*}

\noindent\textbf{Visual MLR:} 
In text MLR, since Eq.~\ref{equ::rank} includes only the base categories in the training set, the resulting learned text embeddings have weak zero-shot recognition of novel categories. To address this issue, we propose the visual MLR branch, which distills knowledge from the CLIP image encoder to achieve better zero-shot recognition performance. This is because the image encoder of CLIP has been trained on a huge amount of images using corresponding text supervision. This enables it to extract more generalized embeddings of images that include features of both base and novel categories, as well as important background environments. To distill these generalized embeddings, similar to text MLR, we use another linear layer $f(\cdot)$ to transform the global image embedding into the learned image embedding $e^{image} = f(e^{\mathit{global}}) \in \mathbb{R}^{d}$. Then, we minimize the distance between $e^{image}$ and the embedding from the CLIP image encoder using a $\mathcal{L}_1$ loss: 
\begin{equation}
  L_{dist} = \sum\limits_{i} \parallel \mathit{IE}(i) - e^{image}_i \parallel_{1},
\end{equation}
where $\mathit{IE}(i)$ denotes the embedding of image $i$ from the CLIP image encoder. Since $\mathit{IE}(i)$ contains rich knowledge of novel categories and important contextual features, the learned $e^{image}$ can well complement $e^{text}$ in text MLR in detecting novel object categories, especially objects whose detection requires the recognition of the global image contexts. During inference, the classification score of a test image $j$ to a category $c$ can be computed as follows:
\begin{equation}\label{eq:visualmlr}
    s_{j,c}^{image} = \mathit{cos}(e_{j}^{image}, t_{c}).
\end{equation}
Since there can still exist a performance gap between visual MLR (student) and CLIP $IE$ (teacher), directly using the CLIP image embedding $IE(j)$ to replace $e_{j}^{image}$ for obtaining ${s}_{j,c}^{image}$ in Eq. \ref{eq:visualmlr} often enables better performance. This variant is denoted as \textbf{visual MLR}$^{+}$. Notably, we adopt the ViT-B/32 CLIP image encoder which takes one $224\times224$ resolution image as input during testing,  so it incurs only minor computation and memory overhead.

\noindent\textbf{Multi-modal MLR:} As discussed above, text MLR and visual MLR branches complement each other in identifying base and novel categories, so we combine their classification scores to have a multi-modal MLR. Particularly, the projection function $f(\cdot)$ in each branch is trained independently. After that, we obtain the learned text embedding $e_{i}^{text}$ and image embedding $e_{i}^{image}$ for each image $i$. Inspired by \cite{gu2021open, kuo2022f},  we ensemble the two probability scores of an image $j$ as follows:
\begin{equation}\label{eqn:mmlr}
  p_{j,c}^{mmlr} = \begin{cases}
    (p_{j,c}^{text})^{\lambda_B} \cdot (p_{j,c}^{image})^{1 - \lambda_B}, & c \in \mathcal{C}_B \\
    (p_{j,c}^{text})^{1 - \lambda_N} \cdot (p_{j,c}^{image})^{\lambda_N}, & c \in \mathcal{C}_N, \\
  \end{cases}
\end{equation}
where $p_{j,c}^{text} = \mathit{sigmoid}(\tilde{s}_{j,c}^{text})$ is a sigmoid classification probability based on a normalized text embedding similarity score $\tilde{s}_{j,c}^{text}=\frac{{s}_{j,c}^{text}-\mu({s}_{j,\cdot}^{text})}{\sigma({s}_{j,\cdot}^{text})}$, and similarly, $p_{j,c}^{image}=\mathit{sigmoid}(\tilde{s}_{j,c}^{image})$ is a classification probability using the image embedding similarity score $s_{j,c}^{image}$ normalized in a similar way as $\tilde{s}_{j,c}^{text}$ ($s_{j,c}^{image}$ can be obtained from visual MLR or visual MLR$^+$), and $\lambda_B$ and $\lambda_N$ is the hyperparameters for controlling the combination of two scores. The zero-mean normalization is used based on the fact that the sigmoid function has a max gradient near zero, so that probability values of positive and negative categories can be better separated, which is beneficial for the detection score refinement process.

\subsection{Context-aware OVOD with Image-Level Multi-modal MLR Scores\label{sec::dsa}}
Our multi-modal MLR learns to recognize both novel and base objects from the global perspective. We show that it can be plugged into different existing OVOD models that are often focused on regional visual concepts via a post-process step to enhance their detection performance. As shown in Fig. \ref{fig:overview}(b), given a test image $\mathbf{I} \in \mathbb{R}^{H \times W \times 3}$, the OVOD model produces a set of instance predictions $\{(b,p^{ovod})_{j}\}$ where $b_j$ is the bounding box coordinates and $p^{ovod}_{j} = \{p^{ovod}_{j, 1}, p^{ovod}_{j, 2},..., p^{ovod}_{j, C}\}$ represents the classification probability scores of the $j$-th instance for all $C=|\mathcal{C}_{\mathit{test}}|$ categories. Then, our MLR model predicts the image-level classification scores $p^{mmlr} = \{p^{mmlr}_{1}, p^{mmlr}_{2},..., p^{mmlr}_{C}\}$. Although the MLR model can not localize the objects, it provides scene context information and prior knowledge about the types of objects that may exist in the whole image. This contextual information enhances the region-level detection performance, especially in detecting the aforementioned hard objects. Therefore, we utilize the following weighted geometric mean to combine the image-level score $p^{mmlr}$ and instance-level score $p^{ovod}_{j}$ as follows:
\begin{equation} \label{equ::dsa}
  p_{j}^{\mathit{cads}} = (p^{mmlr})^{\gamma} \odot (p^{ovod}_{j})^{1 - \gamma},
\end{equation}
where $p_{j}^{\mathit{cads}}$ denotes the context-aware detection score of the $j$-th instance, $\odot$ denotes the element-wise multiplication, and $\gamma$ is a hyperparameter to balance the two types of scores.

\section{Experiments}
\subsection{Datasets}
We evaluate our method on LVIS v1.0 \cite{gupta2019lvis} and COCO \cite{lin2014microsoft} under the open-vocabulary settings, as defined by recent works~\cite{zareian2021open,gu2021open}, with the benchmarks named as OV-COCO and OV-LVIS respectively. 

\noindent\textbf{OV-LVIS:} 
LVIS is a large-vocabulary instance segmentation dataset containing 1,203 categories. The categories are divided into three groups based on their appearance frequency in the dataset: frequent, common, and rare. Following the protocol introduced by \cite{gu2021open}, we treat the frequent and common categories as base categories (noted as LVIS-Base) to train our model. It considers 337 rare categories as novel categories during testing. We report the instance segmentation mask-based average precision metrics for rare (novel), common, frequent, and all classes, denoted as AP$_{r}$, AP$_{c}$, AP$_{r}$, and AP respectively. 

\begin{table*}
  \begin{minipage}{0.68\linewidth}
    \centering
    \caption{Results of different variants of our MLR module on OV-LVIS. }
    \label{variants-table}
    \adjustbox{width=1.0\textwidth}{
      \begin{tabular}{l|ccc|cc|cc}
        \toprule
        Method & \cellcolor{gray!20} AP$_{r}$&AP$_{c}$&AP$_{f}$&R$^{mlr}_{novel}$ &R$^{mlr}_{base}$& GFLOPs & FPS   \\\midrule
        Base Model (BoxSup) & \cellcolor{gray!20}16.4&31.0&35.4&-&- & 215&4.8 \\\midrule
        w/ Text MLR& 
        \cellcolor{gray!20}17.6\textcolor{blue}{(+1.2)} & 31.5\textcolor{blue}{(+0.5)} & 35.5\textcolor{blue}{(+0.1)}&9.7 &65.0 & 233 & 4.6\\
        w/ Visual MLR&\cellcolor{gray!20}20.1\textcolor{blue}{(+3.7)}&31.1\textcolor{blue}{(+0.1)}&34.7\textcolor{red}{(-0.7)}&19.6&32.6 &233 &4.6 \\
        w/ Multi-modal MLR &\cellcolor{gray!20}19.9\textcolor{blue}{(+3.5)}& 31.7\textcolor{blue}{(+0.7)}&35.4\textcolor{gray}{(+0.0)}&21.1&56.8& 233 &4.6 \\
        w/ Visual MLR$^+$&\cellcolor{gray!20}20.6\textcolor{blue}{(+4.2)} & 31.3\textcolor{blue}{(+0.3)} & 34.6\textcolor{red}{(-0.8)}&38.1&31.9 &238&4.5 \\
        
        w/ Multi-modal MLR$^{+}$ &\cellcolor{gray!20}20.3\textcolor{blue}{(+3.9)}&31.8\textcolor{blue}{(+0.8)}&35.5\textcolor{blue}{(+0.1)}&34.4&56.5 & 238 &4.5 \\\hline
        \multirow{2}*{\shortstack{w/ Multi-modal MLR$^{+}$ \\ (ViT-L/14 CLIP)}} &\cellcolor{gray!20} &\multirow{2}*{32.0\textcolor{blue}{(+1.0)}} &\multirow{2}*{35.5\textcolor{blue}{(+0.1)}}&\multirow{2}*{37.8}&\multirow{2}*{58.5}& \multirow{2}*{348} & \multirow{2}*{3.9}  \\
        &\multirow{-2}{*}{\cellcolor{gray!20}21.9\textcolor{blue}{(+5.5)}}&&&&\\\bottomrule
      \end{tabular}}
  \end{minipage}
  \hfill
  \begin{minipage}{0.26\linewidth}
    \centering
    \caption{Effectiveness of hyperparameter $\lambda_B$ and $\lambda_N$ on OV-LVIS. }
    \label{tab::lambda}
    \adjustbox{width=1.0\textwidth}{
    \begin{tabular}{cc|cccc}\toprule
          $\lambda_B$ & $\lambda_N$ & AP$_{r}$&AP$_{c}$&AP$_{f}$&AP\\\midrule
          0.8&1.0&20.3&31.7&35.4&31.1 \\ 
          0.8&0.8&20.3&31.8&35.5&31.3 \\
          0.8&0.6&19.9& 31.8 & 35.5 & 31.2 \\ 
          0.8&0.5&19.1& 31.8 & 35.5 & 30.8 \\  \midrule
          1.0&0.8&20.2& 31.5 &35.5&31.1\\ 
          0.6&0.8&20.3& 31.9 & 35.3 & 31.2 \\ 
          0.5&0.8&20.3& 31.6 & 35.0 & 30.8 \\ 
          \bottomrule
          \end{tabular}
    }
  \end{minipage}
\end{table*}

\noindent\textbf{OV-COCO:} We follow the open-vocabulary setting defined by~\cite{zareian2021open} and split the categories into 48 base categories and 17 novel categories. Only base categories in COCO \texttt{train2017} (noted as COCO-Base) are used for training. We report the box mAP of novel, base, and all classes measured at IoU threshold 0.5, denoted as AP$_{novel}$, AP$_{base}$, and AP respectively. 

\noindent\textbf{Cross-dataset Generalization:} To validate the generalization performance, we train the proposed model on the training set of LVIS and evaluate on two other datasets, Objects365 \cite{shao2019objects365} and OpenImages \cite{kuznetsova2020open}.  We report the AP$_{50}$ result evaluated at IoU threshold 0.5 for both datasets.

\subsection{Implementation Details}
\noindent\textbf{Network Architecture:} We use ResNet-50 \cite{he2016deep} with FPN \cite{lin2017feature} as the default backbone network. To ensure a fair comparison with previous OVOD methods, we adopt the same VLM model, the ViT-B/32 CLIP model, as the teacher model for our MLR module by default. 

\noindent\textbf{Hyperparameters:} The hyperparameters $\lambda_B$, $\lambda_N$, and $\gamma$ are set to 0.8, 0.8, 0.5 for OV-LVIS and 0.8, 0.5, 0.7 for OV-COCO based on the ablation results. 

\noindent\textbf{Training Pipeline:} We adopt the offline training strategy that trains our MLR module separately from the OVOD models. We use AdamW~\cite{loshchilov2017decoupled} optimizer with an initial learning rate of 0.0002 to train our MLR module. For OV-LVIS, our MLR model is trained using the image-level labels of LVIS-Base for 90,000 iterations with a batch size of 64 (48 epochs). As for OV-COCO, we train our model for 12 epochs for a fair comparison with previous OVOD models.   Note that some OVOD models may leverage extra image-level labeled datasets, \eg, ImageNet-21k~\cite{deng2009imagenet}, COCO Caption~\cite{chen2015microsoft}. Therefore, we also train our MLR model using the same dataset for the same iterations when plugging our method into these methods. We use images of size 480×480, augmented with random resized cropping and horizontal flipping during training. 

\noindent\textbf{Inference:} During inference, we use the trained multi-modal MLR module to predict the image-level multi-label scores. We then combine the MLR scores with the instance predictions of the trained OVOD models to obtain the final results based on Eq. \ref{equ::dsa}. Since visual MLR$^+$ generally performs better than its primary version, it is used by default to compute $p_{j,c}^{image}$ in Eq. \ref{eqn:mmlr} during inference in our experiments, denoted as \textbf{multi-modal MLR$^+$}. We show the full results of both versions in Tab.~\ref{variants-table} and our appendix.

\subsection{Main Results}\label{subsec:results}
To evaluate the overall performance and flexibility of our method \textbf{SIC-CADS}, we combine it with various OVOD models that use different strategies, including knowledge distillation (ViLD~\cite{gu2021open}, OC-OVD~\cite{bangalath2022bridging}); prompt learning (POMP~\cite{ren2023prompt}, CORA~\cite{wu2023cora}); self-training (Detic~\cite{zhou2022detecting}); and region-text pre-training (RegionCLIP~\cite{zhong2022regionclip}). The results of these original OVOD models and our SIC-CADS-enabled versions on OV-LVIS, OV-COCO, and cross-dataset generalization are presented below. 

\noindent\textbf{OV-LVIS:} The results of SIC-CADS combined with different OVOD models on OV-LVIS are shown in Tab.~\ref{tab::ov_lvis}. Our method SIC-CADS can consistently and significantly improve all five OVOD models in detecting novel categories, achieving maximal 3.9 gains and setting a new SOTA performance of 26.6 in AP$_r$ when plugged into POMP. Further, SIC-CADS also consistently improves the detection performance of common base categories in AP$_c$, having 0.5-0.9 gains across the models. It retains similar performance in detecting frequent categories in AP$_f$ (increase/decrease in the $[-0.1, 0.2]$ range). These results demonstrate that our proposed multi-modal MLR module effectively learns important global knowledge from CLIP that is complementary to the current regional OVOD models in detecting objects of both novel and base categories. Impressively, this improvement holds for the recent best-performing OVOD models, regardless of whether they exploit external supervision. 

\noindent\textbf{OV-COCO:} SIC-CADS shows similar enabling superiority on the OV-COCO benchmark, as reported in Tab.~\ref{tab::ov_coco}. It is remarkable that SIC-CADS consistently enhances all five SOTA OVOD models in all three metrics, AP$_{novel}$, AP$_{base}$, and AP. Particularly, it increases the AP$_{novel}$ scores by 1.7-3.2 points and AP$_{base}$ by 0.1-1.3 points over the five base models. Notably, RegionCLIP and CORA actually adopt stronger ResNet-50$\times$4 CLIP image encoder as the backbone, and our method still improves their overall performance without using such a strong and large backbone. 

\noindent\textbf{Cross-dataset Generalization:} In the cross-dataset generalization experiment, we train our MLR module using image-level labels of LVIS and combine it with BoxSup and Detic to evaluate the performance on Objects365 and OpenImages. As presented in Tab.~\ref{tab::cross_dataset}, SIC-CADS largely increases AP$_{50}$ by 1.9-2.1 points on Objects365 and 1.5-3.9 points on OpenImages. These results demonstrate our excellent improvement in the generalization ability from the cross-dataset aspect.

\subsection{Further Analysis of SIC-CADS}\label{sec::ablation}

\noindent\textbf{Ablation Study:} Different MLR variants can be used in our SIC-CADS approach. We evaluate the use of different MLR variants on top of the baseline method BoxSup to demonstrate the importance of multi-modal MLR. In addition to the above AP evaluation metrics, We also adopt the recall rate for novel and base categories of the top-10 MLR predictions as an auxiliary metric to evaluate the performance of the MLR module (denoted as R$^{mlr}_{novel}$ and R$^{mlr}_{base}$ respectively). The ablation study is shown in Tab.~\ref{variants-table}. Overall, Text MLR can slightly boost both AP$_r$ and AP$_c$, but it still biases toward the base categories, resulting in very limited gains for novel categories. Visual MLR and Visual MLR$^+$ can significantly increase AP$_r$ by 3.7 - 4.2 points, but AP$_f$ drops by 0.7 - 0.8 points, showing strong zero-shot recognition but relatively weak base category recognition. Multi-modal MLR and Multi-modal MLR$^{+}$ combine the strengths of the two branches and obtain the best recall rate in the base or novel categories, and as a result, they achieve equally excellent results for novel and base category detection, yielding 3.5 - 3.9 gains for AP$_r$ and 0.7 - 0.8 gains for AP$_c$, without harming AP$_f$. Additionally, we also show that using the stronger VLM models, \ie, ViT-L/14 CLIP, can obtain even better performances, achieving up to 5.5 gains for AP$_r$. We also provide the results of using large MLR models (\eg, BLIP\cite{li2022blip} and RAM\cite{zhang2023recognize}) in our appendix.

\noindent\textbf{Computational Costs:} In Tab.\ref{variants-table}, we also show the additional computational overload yielded from our plugged-in module. Generally, our model incurs small additional inference costs when plugged into current OVOD models, resulting in an slight increase in GFLOPs from 215 to 233-238, and a decrease in FPS from 4.8 to 4.6-4.5. This can be attributed to two reasons: (1) we adopt 400×400 resolution during inference since we only need to obtain the image-level labels, while the OVOD model uses 800×800 resolution for more precise box prediction, so we only require approximately 25$\%$ computational cost using the same ResNet50 backbone; (2) downstream tasks (\eg, box regression, instance classification, and non-maximum suppression) of object detectors can result in significant computation, time, and GPU memory costs, which are often 1-3 times as the backbone's cost, while our MLR model simply merges the scores of two branches, which requires minimal costs apart from the backbone's cost. We provide more detailed information in our appendix. 

\begin{figure}[t]
        \centering
        \includegraphics[width=0.45\textwidth]{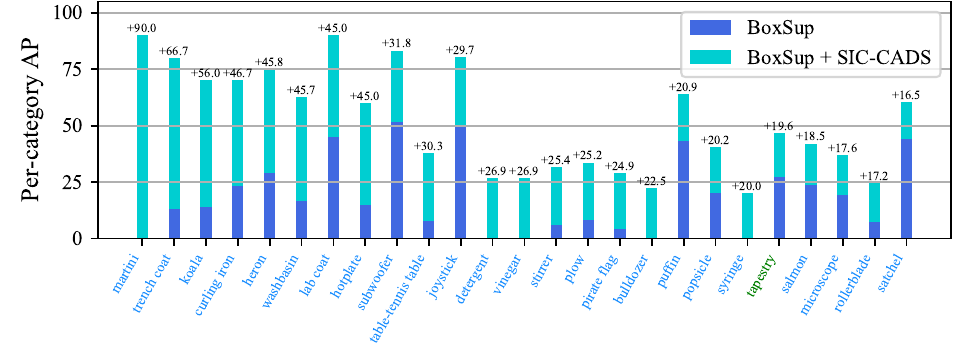}
        \caption{\textbf{Class-specific AP of the top 25 categories with the largest improvement on OV-LVIS.}}\label{fig:per_cat_ap}     
\end{figure}

\noindent\textbf{Analysis of Hyperparameters:} Tab. \ref{tab::gamma} shows the results of varying values of $\gamma$ in Eq.~\ref{equ::dsa} when plugged into BoxSup on OV-LVIS. $\gamma$ controls the combination of the MLR scores and the detection scores. When using a smaller $\gamma$, the MLR scores have only limited impact on the instance scores, leading to smaller gains for AP$_r$. However, if a large $\gamma$ is used, \eg, 0.8, all detection scores will be very close to the MLR score, making all metrics drop sharply. Since $\gamma = 0.5$ achieves a good trade-off between novel and base category detection and obtains the best overall AP, 
we choose it as the default hyperparameter for OV-LVIS. Besides, Tab.~\ref{tab::lambda} shows the results of varying values of $\lambda_B$ and $\lambda_N$ in Eq.~\ref{eqn:mmlr} on LVIS. We find that our method is not sensitive when varying $\lambda_B$ and $\lambda_N$ within the range of [0.8, 1.0]. However, AP$_f$ drops by 0.5 when $\lambda_B = 0.5$, and AP$_r$ drops by 1.2 when $\lambda_N = 0.5$. So we choose $\lambda_B = 0.8$ and $\lambda_N = 0.8$ for OV-LVIS. The ablation results of $\gamma$, $\lambda_B$, and $\lambda_N$ for OV-COCO are shown in our appendix.

\begin{figure}[t]
        \centering
        \includegraphics[width=0.45\textwidth]{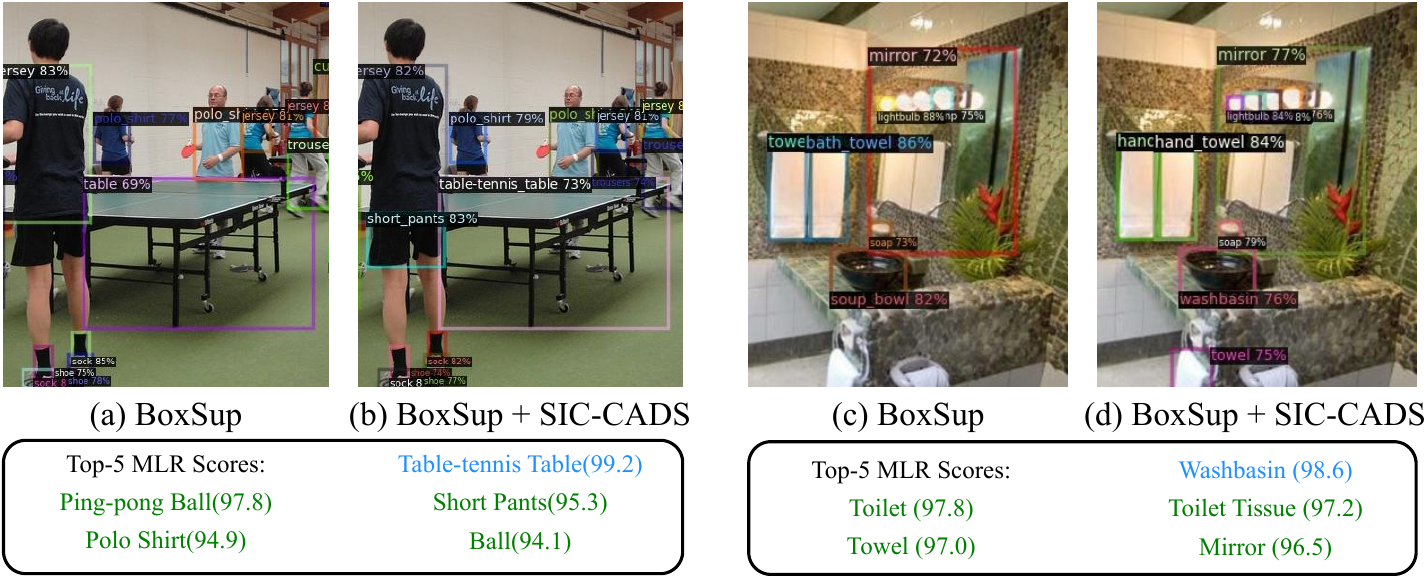}
        \caption{\textbf{Visualizations results of combining SIC-CADS with BoxSup on OV-LVIS.}}
        \label{fig:visualization}
\end{figure}

\noindent\textbf{Qualitative Analysis:}
To further investigate the impact of our method on different object categories, we present a qualitative analysis of the class-specific AP of the top 25 categories with the largest improvement when applying SIC-CADS to BoxSup. The results are illustrated in Fig. \ref{fig:per_cat_ap}, together with detection examples for some of those categories in Fig. \ref{fig:visualization}, where novel and base categories are in \textcolor{cyan! 80}{{blue}} and \textcolor{green!70!black}{green}, respectively. Our proposed method exhibits significant improvement (maximally 90.0 gains in the class-specific AP) in detecting small-sized, blurred, or occluded objects, such as those from the martini, curling iron, washbasin, and table-tennis table categories. These categories are often ambiguous for the regional OVOD models like BoxSup, but our method can effectively detect them due to the contextual knowledge offered by our multi-modal MLR. As also shown in Fig. \ref{fig:visualization}(a-b), SIC-CADS can help recognize fine-grained categories, which are otherwise recognized as coarse-grained categories, \eg, `table' in (a) vs `table-tennis table' in (b). Due to the learned contextual knowledge, SIC-CADS also helps correct wrong detection, \eg, `Soup Bowl' in (c) corrected as `Washbasin' in (d).

\section{Conclusion and Future Work}
This paper proposes SIC-CADS, a novel approach for open-vocabulary object detection that leverages global scene understanding capabilities of VLMs for generalized base and novel category detection. The core in SIC-CADS is a multi-modal MLR module that enables the recognition of different types of objects based on their contextual co-occurrence relations. The resulting MLR scores help largely refine the instance-level detection scores yielded by different types of current SOTA OVOD models that are focused on regional visual concept recognition, enabling significantly improved OVOD performance. This is supported by extensive empirical results on the OV-LVIS, OV-COCO, and the cross-dataset benchmarks: Objects365 and OpenImages. Our qualitative analysis shows that one main source of the significant improvement gained by SIC-CADS is its superior performance in detecting ambiguous novel categories, on which current OVOD models fail to work well.
Despite the promising results, there are still limitations in our method. Particularly, our method may fail when context information does not match the object categories. We discuss such failure cases in our appendix and will improve the method in future work. 

\section{Acknowledgments}
In this work, R. Fang and X. Bai are supported by the National Natural Science Foundation of China 62306247, 62372029. 

\bibliography{aaai24}


\begin{table*}[t]
  \begin{minipage}{0.68\linewidth}
  \caption{Full results on the OV-LVIS benchmark. }\label{tab::full_ov_lvis}
  \centering
  \begin{tabular}{{l|>{\columncolor{gray!20}}cccc}}
    \toprule
    Method &AP$_{r}$&AP$_{c}$&AP$_{f}$ & AP   \\\midrule
    BoxSup~\cite{zhou2022detecting}&16.4&31.0&35.4&30.2 \\
    w/ multi-modal MLR &19.9\textcolor{blue}{(+3.5)}&31.7\textcolor{blue}{(+0.7)}&35.4\textcolor{gray}{(+0.0)}&31.1\textcolor{blue}{(+0.9)} \\ 
    w/ multi-modal MLR$^+$ &20.3\textcolor{blue}{(+3.9)}&31.8\textcolor{blue}{(+0.8)}&35.5\textcolor{blue}{(+0.1)}&31.2\textcolor{blue}{(+1.0)} \\ \midrule

    ViLD~\cite{zhou2022detecting}&16.8&25.6&28.5&25.2 \\
    w/ multi-modal MLR &18.6\textcolor{blue}{(+2.0)}& 26.4\textcolor{blue}{(+0.8)}&28.7\textcolor{blue}{(+0.2)}&26.1\textcolor{blue}{(+0.9)} \\
    
    w/ multi-modal MLR$^+$ &18.7\textcolor{blue}{(+2.1)}& 26.4\textcolor{blue}{(+0.8)}&28.7\textcolor{blue}{(+0.2)}&26.1\textcolor{blue}{(+0.9)} \\ \midrule
    
    RegionCLIP~\cite{zhong2022regionclip} &19.7&28.2&30.7&27.7 \\
    w/ multi-modal MLR &21.6\textcolor{blue}{(+1.9)}&29.1\textcolor{blue}{(+0.9)}&30.8\textcolor{blue}{(+0.1)}&28.4\textcolor{blue}{(+0.7)} \\
    w/ multi-modal MLR$^+$ & 21.9\textcolor{blue}{(+2.2)}&29.1\textcolor{blue}{(+0.9)}&30.9\textcolor{blue}{(+0.2)}&28.5\textcolor{blue}{(+0.8)} \\ \midrule
    OC-OVD~\cite{bangalath2022bridging}&21.1&25.0&29.1&25.9 \\
    w/ multi-modal MLR &22.8\textcolor{blue}{(+1.7)}&25.6\textcolor{blue}{(+0.6)}&29.0\textcolor{red}{(-0.1)}&26.4\textcolor{blue}{(+0.5)} \\ 
    w/ multi-modal MLR$^+$ & 22.5\textcolor{blue}{(+1.4)}&25.6\textcolor{blue}{(+0.6)}&29.0\textcolor{red}{(-0.1)}&26.4\textcolor{blue}{(+0.5)} \\ \midrule
    Detic~\cite{zhou2022detecting}&24.9&32.5&35.6&32.4 \\
    
    w/ multi-modal MLR &26.5\textcolor{blue}{(+1.6)}&32.9\textcolor{blue}{(+0.4)}&35.6\textcolor{gray}{(+0.0)}&32.9\textcolor{blue}{(+0.5)} \\ 
    w/ multi-modal MLR$^+$ & 26.5\textcolor{blue}{(+1.6)}&33.0\textcolor{blue}{(+0.5)}&35.6\textcolor{gray}{(+0.0)}&32.9\textcolor{blue}{(+0.5)} \\ \midrule
    POMP~\cite{ren2023prompt}&25.2&33.0&35.6&32.7 \\
    w/ multi-modal MLR &26.3\textcolor{blue}{(+1.1)}&33.3\textcolor{blue}{(+0.3)}&35.6\textcolor{gray}{(+0.0)}&33.0\textcolor{blue}{(+0.3)} \\ 
    w/ multi-modal MLR$^+$ & 26.6\textcolor{blue}{(+1.4)}&33.3\textcolor{blue}{(+0.3)}&35.6\textcolor{gray}{(+0.0)}&33.1\textcolor{blue}{(+0.4)} \\ \bottomrule
  \end{tabular}
  \end{minipage}
  \hfill
  \begin{minipage}{0.25\linewidth}
    \centering
    \caption{Effectiveness of hyperparameter $\gamma$. }
    \label{tab::gamma}
    \adjustbox{width=1.0\textwidth}{
        \begin{tabular}{c|ccc}\toprule
        $\gamma$ & AP$_{novel}$&AP$_{base}$&AP \\\midrule
        0.3&28.1&51.3 &45.3 \\
        0.4&28.7&52.0 &45.9 \\ 
        0.5&30.1&52.4 &46.6 \\
        0.6&30.7&52.5 &46.7 \\ 
        0.7&31.0&52.4 &46.8 \\ 
        0.8&30.5&51.8 &46.2 \\ 
        0.9&30.2&49.1 &44.1 \\ \bottomrule
        \end{tabular}
    }
  \end{minipage}
\end{table*}

\section{Supplementary Material}
We present additional experimental and qualitative results in this supplementary material. 
\section{Additional Experimental Results}
\subsection{Full Results on OV-LVIS and OV-COCO}
Our main paper only shows the main results using our best-performed multi-modal MLR$^+$ module, which directly uses the CLIP Image Encoder (teacher) during inference. This section reports the full results of using two different MLR variants on the OV-LVIS and OV-COCO benchmarks, including both multi-modal MLR and multi-modal MLR$^+$. As shown in Tab.~\ref{tab::full_ov_lvis}, although there exists a performance gap between our Visual MLR (student) and CLIP Image Encoder (teacher), we can still achieve consistent improvement for these SOTA OVOD models on OV-LVIS, yielding 1.1-3.5 gains for AP$_r$. Notably, as shown in Tab.~\ref{tab::full_ov_coco}, multi-modal MLR can maintain performance comparable to multi-modal MLR$^+$ on OV-COCO. The difference in performance on the LVIS and COCO datasets may be because the distilled knowledge is sufficient to deal with the relatively coarse-grained classification of COCO, but the large-vocabulary classification of LVIS still requires the powerful zero-shot recognition ability of the CLIP Image Encoder for better OVOD performance. 

\subsection{Ablation Study of Hyperparameters on OV-COCO}
The main paper presents the hyperparameter analysis results on OV-LVIS. Here Tab. \ref{tab::gamma} shows the results of varying values of $\gamma$ in Eq. 7 when plugged into Detic on OV-COCO. When using a smaller $\gamma$ (\eg, 0.3 - 0.4), the MLR scores have only limited impact on the instance scores, leading to smaller gains for AP$_{novel}$. However, if a large $\gamma$ is used, \eg, 0.9, all detection scores will be very close to the MLR score, making all metrics drop sharply. We find that $\gamma = 0.7$ achieves the best overall detection performance, so we choose it as the default hyperparameter for OV-COCO. Besides, Tab. \ref{tab::lambda} shows the results of varying values of $\lambda_B$ and $\lambda_N$ in Eq. 6 on OV-COCO.We find that choosing $\lambda_N = 0.5$ and $\lambda_B = 0.8$ yields the best results. So we choose $\lambda_B = 0.8$ and $\lambda_N = 0.5$ by default for plugging SIC-CADS into different OVOD methods on OV-COCO.

\begin{table}[t]
    \centering
    \caption{Results of w/ or w/o normalization on OV-LVIS.}
    \adjustbox{width=0.4\textwidth}{
    \begin{tabular}{l|cccc}
    \toprule
    Method&AP$_{r}$&AP$_{c}$&AP$_{f}$ & AP \\\midrule
    BoxSup&16.4&31.0&35.4&30.2  \\\midrule
    + SIC-CADS (w/o normalization) &18.0&31.2&35.4&30.6  \\\midrule
    + SIC-CADS (w/ normalization) & 20.3&31.8&35.5&31.2\\\bottomrule
    \end{tabular}
    
    \label{tab:norm}
}
    
\end{table}

\begin{table}[t]
    \centering
    \caption{Results of online training strategy on OV-LVIS. We adopt the multi-modal MLR variant in this table. }
    \adjustbox{width=0.4\textwidth}{
    \begin{tabular}{l|cccc}
    \toprule
    Method&AP$_{r}$&AP$_{c}$&AP$_{f}$ & AP \\\midrule
    BoxSup&16.4&31.0&35.4&30.2  \\\midrule
    + SIC-CADS (online training) &19.2&31.2&35.2&30.6  \\\midrule
    + SIC-CADS (offline training) & 19.9&31.7&35.4&31.0\\\bottomrule
    \end{tabular}
    
    \label{tab:online}
}
    
\end{table}

\begin{table*}[t]
  \begin{minipage}{0.68\linewidth}
  \centering
  \caption{Full results on the OV-COCO benchmark. }\label{tab::full_ov_coco}
  \begin{tabular}{l|>{\columncolor{gray!20}}ccc}
    \toprule
    Method & AP$_{novel}$&AP$_{base}$& AP   \\\midrule
    Detic~\cite{zhou2022detecting}&27.8&51.1&45.0 \\
    w/ multi-modal MLR& 31.0\textcolor{blue}{(+3.2)}&52.4\textcolor{blue}{(+1.3)}&46.8\textcolor{blue}{(+1.7)} \\ 
    w/ multi-modal MLR$^+$& 31.0\textcolor{blue}{(+3.2)}&52.4\textcolor{blue}{(+1.3)}&46.8\textcolor{blue}{(+1.7)} \\ \midrule
    BARON~\cite{wu2023aligning}&35.1&55.2&49.9 \\
    w/ multi-modal MLR& 36.8\textcolor{blue}{(+1.7)}&56.0\textcolor{blue}{(+0.8)}&51.0\textcolor{blue}{(+1.1)} \\ 
    w/ multi-modal MLR$^+$&36.9\textcolor{blue}{(+1.8)}&56.1\textcolor{blue}{(+0.9)}&51.1\textcolor{blue}{(+1.2)} \\ \midrule
    RegionCLIP~\cite{zhong2022regionclip}&39.3&61.6&55.7 \\
    w/ multi-modal MLR &41.1\textcolor{blue}{(+1.9)}&61.6\textcolor{gray}{(+0.0)}&56.1\textcolor{blue}{(+0.4)}\\ 
    w/ multi-modal MLR$^+$ &41.4\textcolor{blue}{(+2.1)}&61.7\textcolor{blue}{(+0.1)}&56.3\textcolor{blue}{(+0.6)}\\ \midrule
    OC-OVD~\cite{bangalath2022bridging}&40.7&54.1&50.6 \\
    w/ multi-modal MLR &43.0\textcolor{blue}{(+2.3)}&55.1\textcolor{blue}{(+1.0)}&51.9\textcolor{blue}{(+1.3)} \\ 
    w/ multi-modal MLR$^+$ &42.8\textcolor{blue}{(+2.1)}&55.1\textcolor{blue}{(+1.0)}&51.9\textcolor{blue}{(+1.3)} \\ \midrule
    CORA~\cite{wu2023cora}&41.6&44.7&43.9 \\
    w/ multi-modal MLR&43.2\textcolor{blue}{(+1.6)}&45.7\textcolor{blue}{(+1.0)}&45.1\textcolor{blue}{(+1.2)} \\ 
    w/ multi-modal MLR$^+$&43.3\textcolor{blue}{(+1.7)}&45.7\textcolor{blue}{(+1.0)}&45.1\textcolor{blue}{(+1.2)} \\ \bottomrule
  \end{tabular}
  \end{minipage}
  \hfill
  \begin{minipage}{0.3\linewidth}
    \centering
    \caption{Effectiveness of hyperparameter $\lambda_B$ and $\lambda_N$ on OV-COCO. }
    \label{tab::lambda}
    \adjustbox{width=1.0\textwidth}{
    \begin{tabular}{cc|ccc}\toprule
          $\lambda_B$ & $\lambda_N$ & AP$_{novel}$&AP$_{base}$&AP\\\midrule
          0.8&0.9&29.3&52.4 &46.4 \\ 
          0.8&0.7&30.1&52.3 &46.6 \\
          0.8&0.5&31.0&52.4 &46.8 \\  
          0.8&0.3&29.6&52.4 &46.4 \\\midrule
          1.0&0.5&30.9&52.5 &46.8 \\ 
          0.6&0.5&31.0&51.8 &46.3  \\ 
          0.5&0.5&31.0&51.5 &46.1 \\ 
          \bottomrule
          \end{tabular}
    }
  \end{minipage}
\end{table*}

\subsection{Effect of the Normalization Operation in Eq. 6}
\begin{figure*}
\tiny
  \centering
  \begin{subfigure}[b]{0.3\textwidth}
    \includegraphics[width=\textwidth]{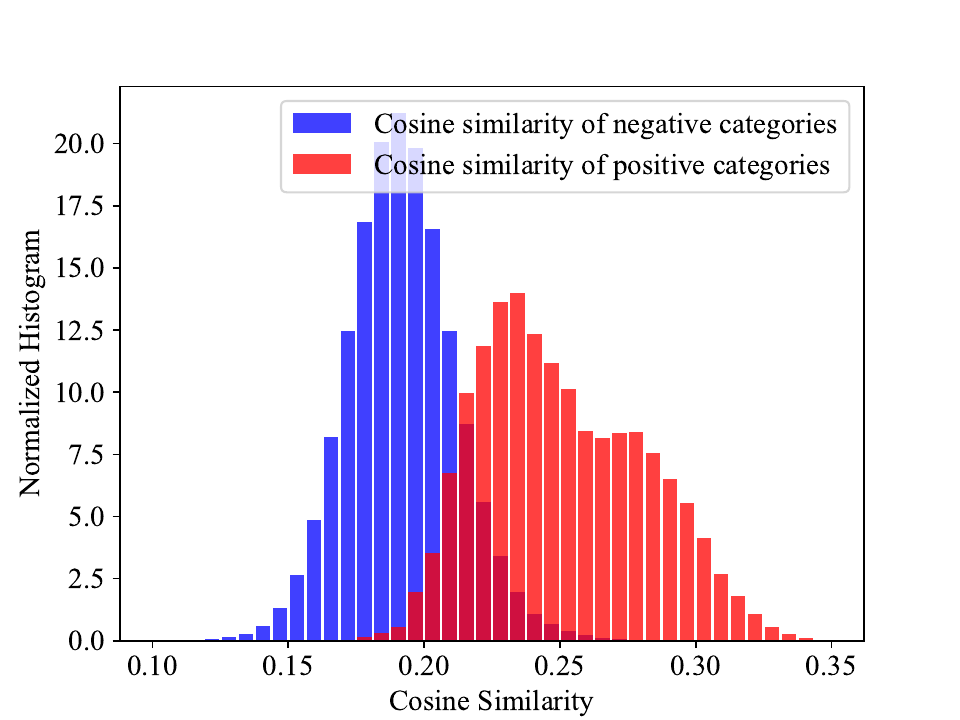}
    \centering
    \caption{Unnormalized similarity scores.}
    \label{fig:subfig1}
  \end{subfigure}
  \begin{subfigure}[b]{0.3\textwidth}
    \includegraphics[width=\textwidth]{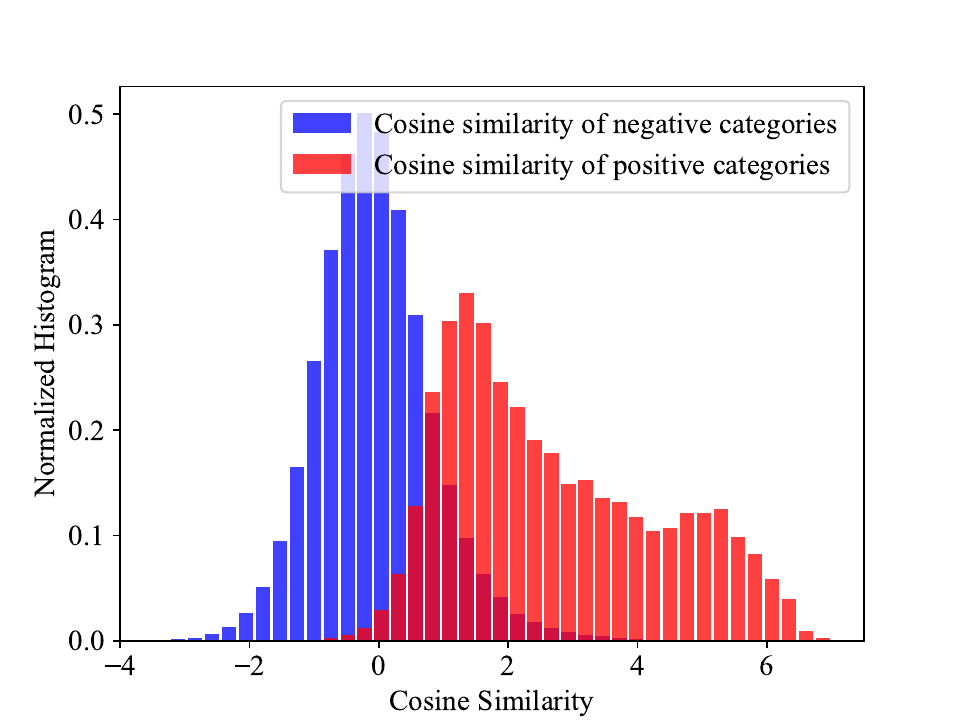}
    \centering
    \caption{Normalized similarity scores.}
    \label{fig:subfig2}
  \end{subfigure}
  \\
  \begin{subfigure}[b]{0.3\textwidth}
    \includegraphics[width=\textwidth]{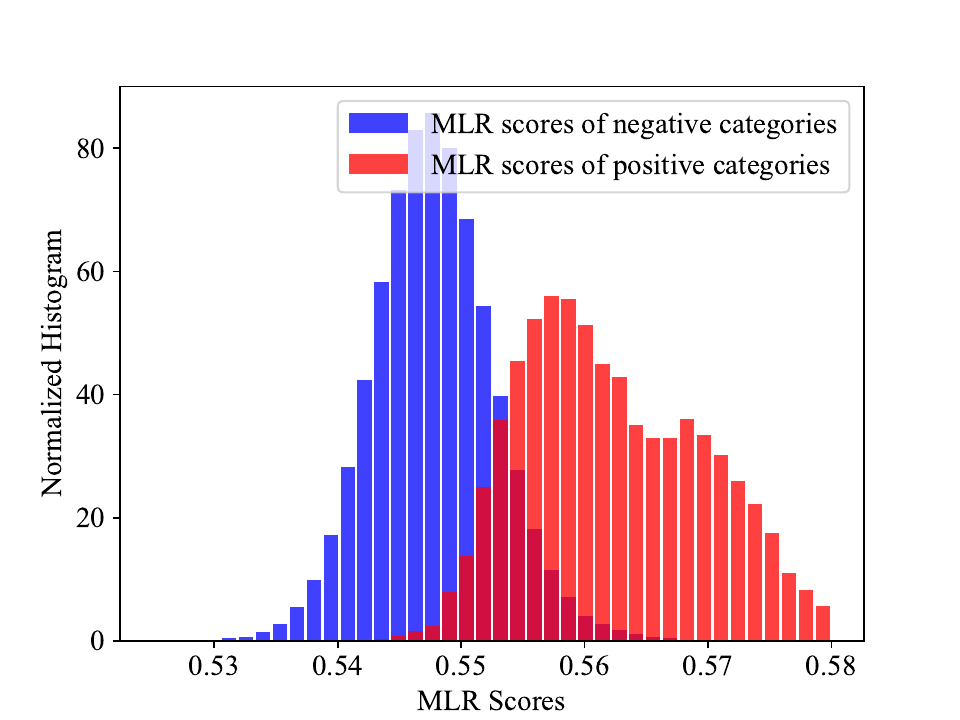}
    \centering
    \caption{Unnormalized MLR Scores.}
    \label{fig:subfig3}
  \end{subfigure}
  \begin{subfigure}[b]{0.3\textwidth}
    \includegraphics[width=\textwidth]{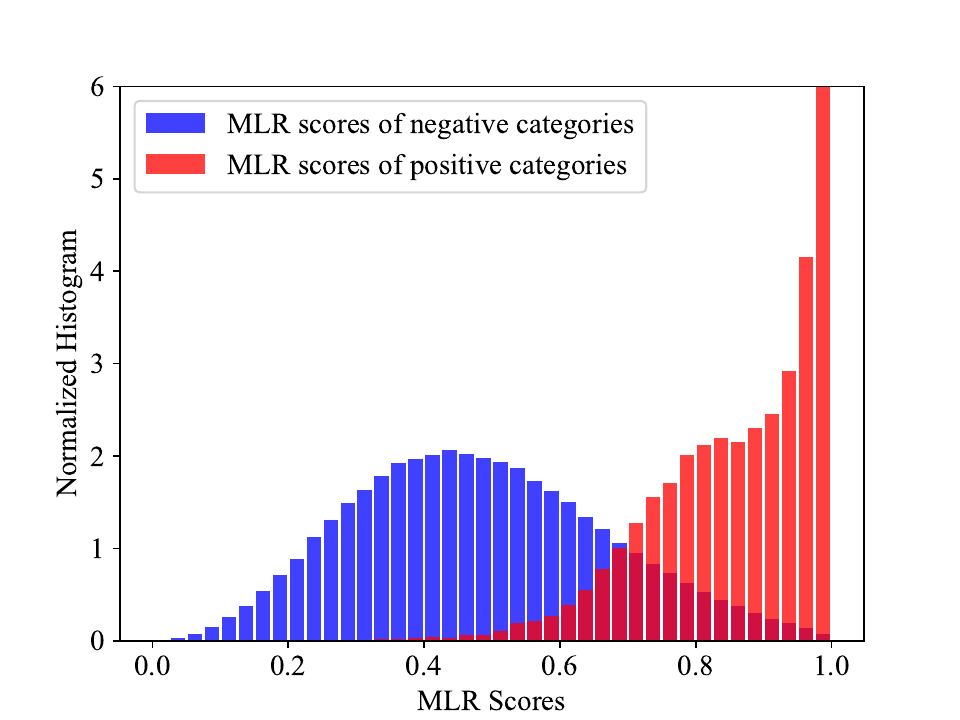}
    \centering
    \caption{Normalized MLR Scores.}
    \label{fig:subfig4}
  \end{subfigure}
  \caption{Effect of using the normalization operation in Eq. 6 on OV-LVIS. (a) and (c) are the resulting unnormalized similarity scores, while (b) and (d) present the similarity scores using our normalization operation. MLR scores are obtained via feeding similarity scores into the sigmoid function.}
  \label{fig:dist}
\end{figure*}

The normalization operation in Eq. 6 is a simple yet effective approach for separating scores of positive categories (\ie, categories that appear in the image) and negative categories (\ie, categories that this image does not contain) during testing. We first obtain the cosine similarity scores by Eq. 2 or Eq. 5 and then compute the mean and standard deviation using scores of all categories in one image. For example, given the similarity scores for the i-th image $s_i = \{s_{i,1}, s_{i,2},...,s_{i,C}\}$, the score of the c-th category is normalized as $s_{i,c}^{normed}=\frac{s_{i,c}-mean(s_{i})}{std({s_i})}$. 
As shown in Fig. \ref{fig:dist}, when we do not apply the score normalization, the similarity scores are distributed in [0.1, 0.35]. When directly feeding them into a Sigmoid function, the MLR scores of positive and negative categories are distributed in [0.53, 0.58]. Such narrowly distributed MLR scores fail to provide effective guidance for detection score refinement in Eq. 7 (\eg, they can not effectively decrease the scores of negative instances and raise the scores of positive instances). When performing the normalization operation, the logits and MLR scores can be widely spread into [-4, 7] and [0, 1], so that the probability scores of negative and positive categories can be clearly separated. As shown by the results in Tab. \ref{tab:norm}, using normalization enables our model to work much better, especially on AP$_r$.

\subsection{Using Large Pre-trained MLR Models}
Since our proposed approach is generic, we can replace our multi-modal MLR model with larger, stronger pre-trained models, such as BLIP \cite{li2022blip} and Recognize Anything Model (RAM) \cite{zhang2023recognize}, to achieve even better novel object detection. As shown in Tab. \ref{tab:large}, using large pre-trained models like BLIP and RAM achieves 5.7-6.3 gains for AP$_r$. However, they also result in a 50.9\% to 63.7\% increase in computational overload. In contrast, our multi-modal MLR model has a very simple, lightweight structure that is easy to reproduce and implement. Thus, our default model is to have a good balance between model performance and computation overload, but its detection accuracy can be further enhanced if more computational cost is allowed.

\begin{table*}
\centering
\caption{Results of using large pre-train MLR models on OV-LVIS. }
\label{tab:large}
\adjustbox{width=0.9\textwidth}{
  \begin{tabular}{l|ccc|cc|cc}
    \toprule
    Method & \cellcolor{gray!20} AP$_{r}$&AP$_{c}$&AP$_{f}$&R$^{mlr}_{novel}$ &R$^{mlr}_{base}$& GFLOPs & Inference Time (Ms/Image)   \\\midrule
    Base Model (BoxSup) & \cellcolor{gray!20}16.4&31.0&35.4&-&- & 215.2&207.5 \\\midrule
    w/ Multi-modal MLR &\cellcolor{gray!20}19.9\textcolor{blue}{(+3.5)}& 31.7\textcolor{blue}{(+0.7)}&35.4\textcolor{gray}{(+0.0)}&21.1&56.8& 233.2(+8.3\%) &218.2(+5.1\%) \\

    \multirow{2}*{\shortstack{w/ Multi-modal MLR$^{+}$ \\ (ViT-B/32 Backbone)}}&\cellcolor{gray!20}&\multirow{2}*{31.8\textcolor{blue}{(+0.8)}}&\multirow{2}*{35.5\textcolor{blue}{(+0.1)}}&\multirow{2}*{34.4}&\multirow{2}*{56.5} &\multirow{2}*{238.4(+10.7\%)}  &\multirow{2}*{224.0(+7.9\%)} \\
    
    &\multirow{-2}{*}{\cellcolor{gray!20}20.3\textcolor{blue}{(+3.9)}}&&&&\\ \midrule
    
    \multirow{2}*{\shortstack{w/ BLIP \cite{li2022blip} \\ (ViT-L/14 Backbone)}} &\cellcolor{gray!20} &\multirow{2}*{31.4\textcolor{blue}{(+0.4)}} &\multirow{2}*{35.1\textcolor{red}{(-0.3)}}&\multirow{2}*{39.5}&\multirow{2}*{46.7}& \multirow{2}*{352.6(+63.7\%)} & \multirow{2}*{256.3(+23.5\%)}  \\
    &\multirow{-2}{*}{\cellcolor{gray!20}22.1\textcolor{blue}{(+5.7)}}&&&&\\\midrule
    
    \multirow{2}*{\shortstack{w/ RAM \cite{zhang2023recognize} \\ (Swin-L Backbone)}} &\cellcolor{gray!20} &\multirow{2}*{31.6\textcolor{blue}{(+0.6)}} &\multirow{2}*{35.3\textcolor{red}{(-0.1)}}&\multirow{2}*{45.6}&\multirow{2}*{50.3}& \multirow{2}*{324.8(+50.9\%)} & \multirow{2}*{243.6(+17.7\%)}  \\
    &\multirow{-2}{*}{\cellcolor{gray!20}22.7\textcolor{blue}{(+6.3)}}&&&&\\ \bottomrule

  \end{tabular}}

\end{table*}

\subsection{Training and Inference Cost of Our Module}
Since our model simply performs the image-level multi-label classification task, the hardware requirements for training are very low, e.g., it can be trained using one single RTX-3090Ti GPU for 22 hours. As for the inference process, our model generally incurs small additional costs when plugged into current OVOD models for two reasons: (1) we adopt 400×400 resolution during inference since we only need to obtain the image-level labels, while the OVOD model uses 800×800 resolution for more precise box prediction, so we only require approximately 25$\%$ computational cost using the same ResNet50 backbone; (2) downstream tasks (\eg, box regression, instance classification, and non-maximum suppression) of object detectors can result in significant computation, time, and GPU memory costs, which are often 1-3 times as the backbone's cost, while our MLR model simply merges the scores of two branches, which requires minimal costs apart from the backbone's cost. For a more detailed analysis, Tab. \ref{tab:cost_mlr} shows the inference costs of only using our MLR model on OV-LVIS, including inference time, memory consumption, and computation cost. Tab. \ref{tab:cost_infer} shows the additional inference costs of combining our MLR module with Detic on OV-LVIS and OV-COCO benchmarks. Since the time costs of the NMS process in 1203-class LVIS are significantly larger than 80-class COCO, our inference time increment percentage is relatively larger on OV-COCO (\eg, 14.7\%-17.8\% on OV-COCO vs 5.1\%-7.9\% on OV-LVIS).

\begin{table*}[t] 
\centering
\caption{Inference cost of our MLR model on OV-LVIS.}\label{tab:cost_mlr}
\begin{tabular}{l|cccc}
\toprule
model&Backbone &Inference Time (Ms/Image)&Memory (Gb)&GFLOPs \\\midrule
multi-modal MLR & ResNet50&11.3&0.7&18.0 \\\midrule
multi-modal MLR$^+$ & ResNet50 + ViT/B32&16.7&1.1&23.2 \\\bottomrule
\end{tabular}

\end{table*}

\subsection{Online Training Strategy}
Since we adopt an offline training strategy that uses separate backbones for MLR module and OVOD model. Although it can be easily combined with trained OVOD models as a plug-in component, it inevitably incurs additional computation costs for the separate backbone network. Therefore, we also try to jointly train both the MLR module and OVOD model using the same backbone to further boost our inference efficiency. As shown in Tab. \ref{tab:online}, AP$_{r}$, AP$_{c}$, and AP$_{f}$ drop by 0.7, 0.5, and 0.2 respectively when we adopt the online training strategy. This is possibly caused by the conflicts between multi-task loss functions. So we still need to handle the multi-task conflict in our future work and further improve the inference efficiency for real-time scenarios.

\section{Additional Qualitative Results}
\subsection{Qualitative Results on OV-COCO}
\begin{figure*}[!t]
        \centering
        \includegraphics[width=1.0\textwidth]{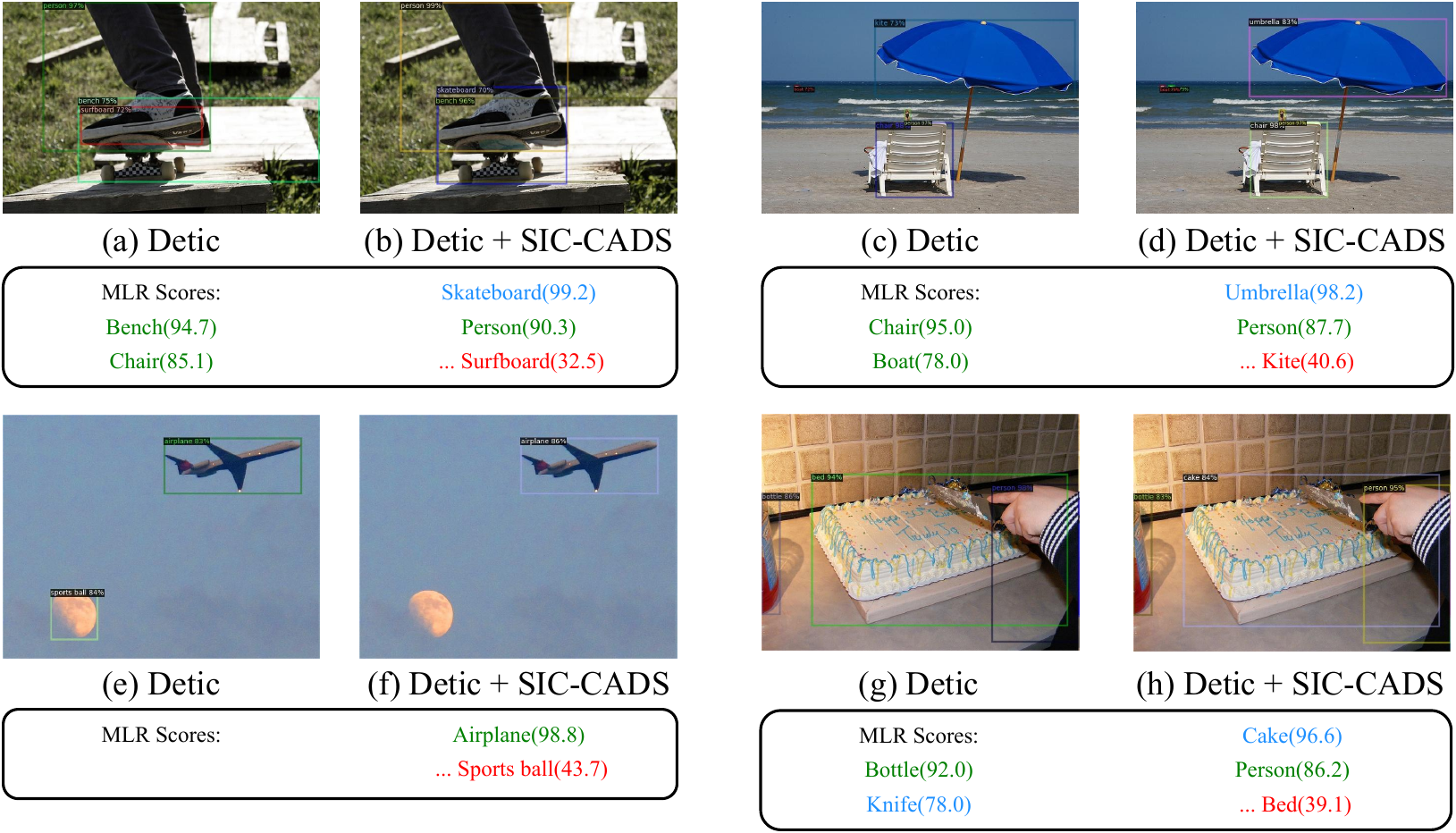}
        \caption{\textbf{Qualitative results on OV-COCO.}}\label{fig:vis_coco}       
\end{figure*}

Fig.~\ref{fig:vis_coco} visualizes the detection results when applying SIC-CADS to Detic on OV-COCO. Our model can correct the wrong object predictions by leveraging the global contextual information, \eg, the misclassified `surfboard' and `kite' in (a) and (c) are respectively corrected as `skateboard' and `umbrella' in (b) and (d) by capturing the background feature of grassland and beach. Similar results can be seen in (e)-(h).

\begin{table*}[t] 
\centering
\caption{Inference cost of combining our MLR model with Detic on OV-LVIS and OV-COCO.}\label{tab:cost_infer}
\begin{tabular}{l|cccccc}
\toprule
model&Dataset&Backbone&Detector &Inference Time (Ms/Image)&Memory (Gb)&GFLOPs \\\midrule
Detic&LVIS&ResNet50&CenterNet-V2&207.5&2.6&215.2 \\\midrule
w/ multi-modal MLR&LVIS&ResNet50&CenterNet-V2&218.2(+5.1\%)&3.0&233.2(+8.3\%) \\\midrule
w/ multi-modal MLR$^+$&LVIS&ResNet50&CenterNet-V2&224.0(+7.9\%)&3.3&238.4(+10.7\%) \\\bottomrule \toprule
Detic&COCO&ResNet50&Faster-RCNN&76.8&2.2&185.4 \\\midrule
w/ multi-modal MLR&COCO&ResNet50&Faster-RCNN&88.1(+14.7\%)&2.6&203.4(+9.7\%) \\\midrule
w/ multi-modal MLR$^+$&COCO&ResNet50&Faster-RCNN&93.5(+17.8\%)&2.9&208.6(+12.5\%) \\\bottomrule 
\end{tabular}

\end{table*}

\subsection{More Failure Cases under Uncommon Situations}
\begin{figure}[h]
        \centering
        \includegraphics[width=0.45\textwidth]{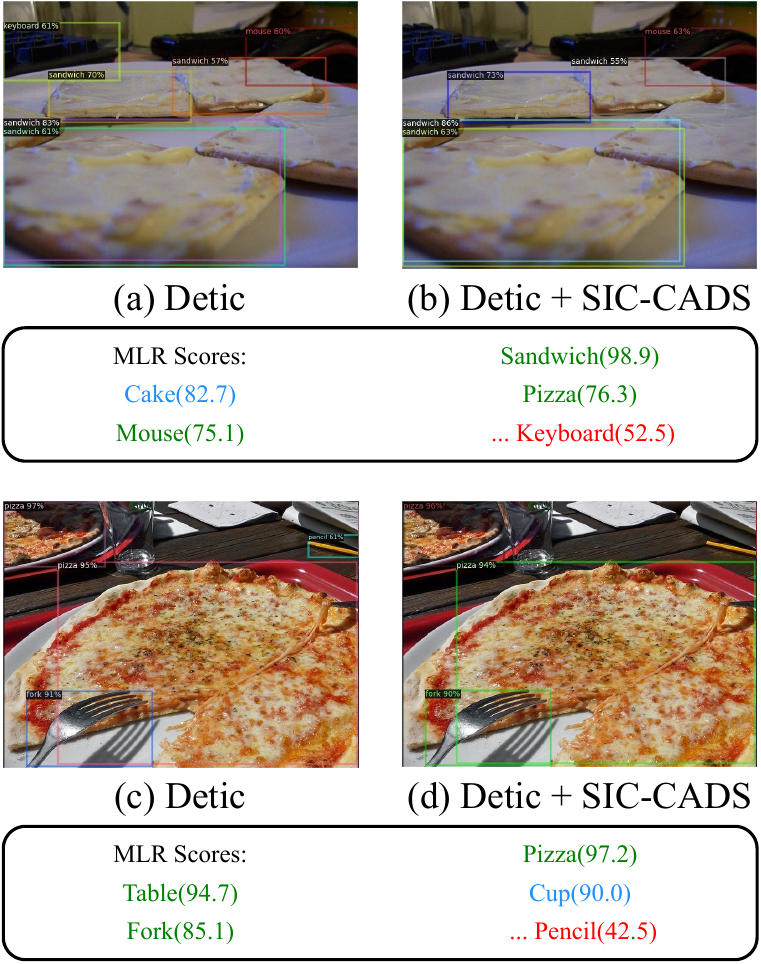}
        \caption{\textbf{Failure cases of our method.}}\label{fig:vis_fail}       
\end{figure}

Although our MLR module can leverage the context feature that benefits the recognition of hard objects. However, when the context information does not match the object category, it may mislead the OVOD models to the wrong predictions. As shown in Fig. \ref{fig:vis_fail}, the MLR scores of `keyboard' and `pencil' are low in (a) and (c), because 
the context information (\eg, `sandwich' and `pizza') is not helpful for the detection of their existence. Therefore, our method leads to the wrong prediction as shown in (b) and (d). However, such uncommon cases are rare in the dataset. For example, we have reviewed the detection results of 5,000 images on OV-COCO but only found less than 10 images with these failure cases. We will try to handle such a difficult problem in our future work.

\end{document}